\documentclass[12pt]{article}

\usepackage[]{algorithmic}
\usepackage[]{algorithm2e}
\usepackage{caption}
\usepackage{float}
\usepackage{graphicx}
\usepackage[%
	colorlinks=true,
	pdfborder={0 0 0},
	linkcolor=blue,
	urlcolor=blue
]{hyperref}
\usepackage{svg}
\usepackage{times}
\usepackage{titlesec}

\titleformat{\paragraph}
{\normalfont\normalsize\bfseries}{\theparagraph}{1em}{}
\titlespacing*{\paragraph}
{0pt}{3.25ex plus 1ex minus .2ex}{1.5ex plus .2ex}

\title{Interaction Networks: Using a Reinforcement Learner to train other Machine Learning algorithms}
\author{Florian Dietz \\ \href{mailto:floriandietz44@gmail.com}{floriandietz44@gmail.com}}

\begin{document}

\maketitle

\begin{abstract}

The wiring of neurons in the brain is more flexible than the wiring of connections in contemporary artificial neural networks. It is possible that this extra flexibility is important for efficient problem solving and learning.

This paper introduces the Interaction Network. Interaction Networks aim to capture some of this extra flexibility.

An Interaction Network consists of a collection of conventional neural networks, a set of memory locations, and a DQN or other reinforcement learner. The DQN decides when each of the neural networks is executed, and on what memory locations. In this way, the individual neural networks can be trained on different data, for different tasks. At the same time, the results of the individual networks influence the decision process of the reinforcement learner. This results in a feedback loop that allows the DQN to perform actions that improve its own decision-making.

Any existing type of neural network can be reproduced in an Interaction Network in its entirety, with only a constant computational overhead.

Interaction Networks can then introduce additional features to improve performance further. These make the algorithm more flexible and general, but at the expense of being harder to train.

In this paper, thought experiments are used to explore how the additional abilities of Interaction Networks could be used to improve various existing types of neural networks.

Several experiments have been run to prove that the concept is sound. These show that the basic idea works, but they also reveal a number of challenges that do not appear in conventional neural networks, which make Interaction Networks very hard to train.

Further research needs to be done to alleviate these issues. A number of promising avenues of research to achieve this are outlined in this paper.


\end{abstract}

\section{Introduction}

Since their inception, many different types of neural networks have been developed, most of them specialized for a specific purpose: Recurrent Neural Networks to deal with sequences, Convolutional Neural Networks to understand images more effectively, Attention Mechanisms to focus on items based on their relevance, Generative Adversarial Networks to reproduce distributions, and many more.

But one underlying aspect of neural architecture has remained unchanged: To solve a problem the Neural Network is always applied to an input to produce an output. The format of the input is fixed. The format of the output is fixed. The forward pass of the network always performs the same steps. The same mathematical operations are applied in the same order every time. The only thing that changes are the weights.

The human brain is a very complex system that does not have this limitation\cite{tononi1998complexity,bullmore2012economy,stam2014modern}. In the brain, information flows are not static. For one example, attention can be focused volitionally by “top-down” signals derived from task demands and automatically by “bottom-up” signals from salient stimuli\cite{buschman2007top}. For another example, the flow of information in the brain is reversed during visual imaginery as compared to visual perception\cite{dentico2014reversal}. At the same time, the brain has dedicated regions and processes for storing and processing semantic memories\cite{martin2001semantic}.

It stands to reason that Artificial Neural Networks could benefit from this flexibility as well.

Expressed in the form of commonly used data structures, the brain can be viewed as a hypergraph structure, where each vertex holds data, while each edge is a neural network that processes data from some vertices and puts the results in some other vertices. These edges / neural networks can be executed in any order, and can even process the data in a circular fashion. Some unknown mechanism controls the order of execution of the edges in the brain. A reinforcement learner seems a reasonable approximation of this control mechanism for the purposes of replicating the abilities of this hypergraph.

\section{Interaction Networks}

This paper introduces the Interaction Network (IN).

The Interaction Network is a generalization of Neural Networks.

Any existing type of neural network can be reproduced in an Interaction Network in its entirety, with only a constant computational overhead.

Interaction Networks can then introduce additional features to improve performance further. These make the algorithm more flexible and general, but at the expense of being harder to train.

\subsection{High-level Overview}

A reinforcement learner is put in charge of managing a collection of other machine learning algorithms. An internal mental state is modeled, which can be altered by the individual learning algorithms, and which is fed back into the Reinforcement Learner as an input.

Interaction Networks have a number of interesting features that are promising candidates for improving upon existing neural networks.

Of course, there is no such thing as a free lunch. It is quite possible that these benefits will be offset by other drawbacks, or simply turn out to be too hard to utilize in practice.

However, the sheer number of additional options provided by Interaction Networks makes it very likely that at least one of them will turn out to be very useful:

\begin{itemize}
	\item Interaction Networks are autonomous agents. They are lifelong-learning systems with a continuous lifetime that interact with their environment as needed, on their own initiative. This is in contrast to normal neural networks, which are essentially functions that map inputs to outputs.
	\item Interaction Networks can learn to solve multiple different types of tasks at the same time. This has previously proven to improve performance even on individual tasks, by allowing the network to generalize better\cite{ruder2017overview}. In a similar vein, Transfer Learning is a natural consequence of the way IN's work, as IN's will automatically try to use previously learned ideas to solve new problems.
	\item Existing neural networks can be embedded in an Interaction Network after being trained separately. The IN will be able to learn to make use of them as needed, It can even combine previously independent algorithms with each other to solve more complex problems.
	\item Interaction Networks can combine different input formats, can choose to keep or discard any data they receive based on salience, and can even actively seek out additional data from external APIs. Combining multiple sources of data has been shown to improve performance on conventional neural networks\cite{kaiser2017one}.
	\item Interaction Networks contain a memory system that can be used both for short term memory and for long-term associative memory. With enough effort, the system can even be trained to use hierarchical knowledge storage and perform symbolic reasoning.
\end{itemize}

\subsection{Definition}

\subsubsection{Structure}

Structurally, an Interaction Network consists of the following components:

\begin{itemize}
	\item The \textbf{Control Unit} (CU), which decides what the IN should do next. This is the DQN in charge of the Interaction Network.
	\item A set of \textbf{Nodes}, which store data in the form of tensors.
	\item A set of \textbf{Processing Units} (PU's). Each Processing Unit is a conventional neural network of arbitrary complexity. A PU takes one or more Nodes as inputs, and one or more Nodes as outputs.
	\item \textbf{Environments}. Environments provide a way for the IN to interact with the outside world. Environments can connect to Nodes in order to provide input to the IN, read output of the IN, and provide feedback in the form of loss functions and gradients to the IN. For advanced usecases, they can also be used to make helper functions and external API's available to the IN.
\end{itemize}

\begin{figure}
	\centering
	\fbox{
		\includegraphics[width=1.0\columnwidth]{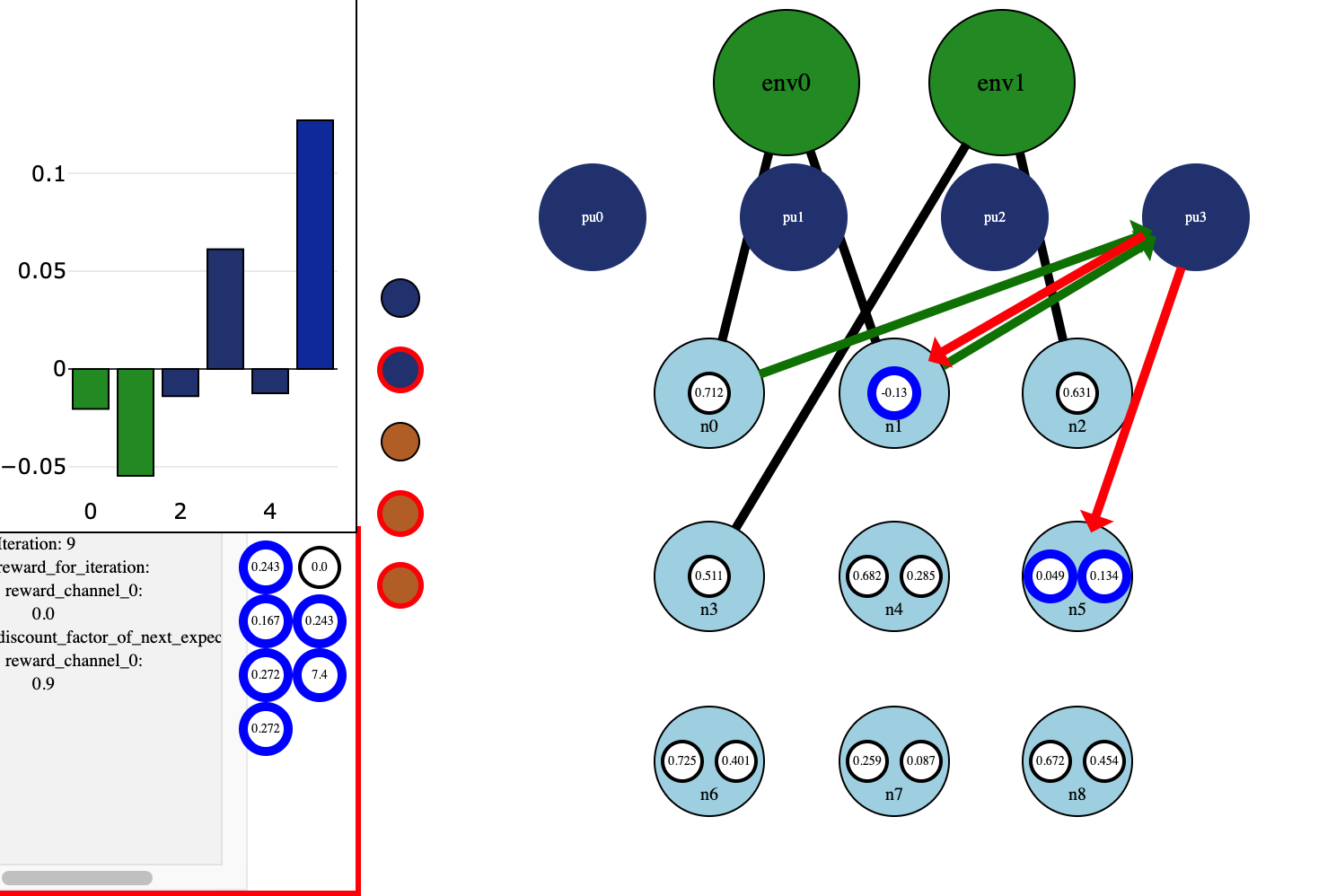}
	}
	\captionsetup{singlelinecheck=off,font=footnotesize}
	\caption[list=off]{This image shows an Interaction Network. Note that this image is a screenshot of an interactive visualization, viewable in Jupyter notebooks. It shows a snapshot of an Interaction Network at a fixed point in time. See below for the code to generate such visualizations.
		\begin{itemize}
			\item The light blue circles represent Nodes. Each white circle with a number in it represents a tensor. Some Nodes have only one tensor, while others have two. The optional second tensor is used as an input by the Control Unit.
			\item The dark blue circle represent Processing Units. In the current iteration the Processing Unit pu3 is being executed. It uses the Nodes n0 and n1 as inputs, and writes the outputs back into n1 and n5.
			\item The green circles represent Environments. They interact with the Nodes they are connected to.
			\item The graph at the top-left shows the current state of the Control Unit. Each bar represents one action that could be taken. One for each of the Environments, and one for each of the Processing Units.
			\item The bottom left displays various information about the current state of the network. For example, we are currently looking at iteration 9.
		\end{itemize}
		You can ignore the numbers in the tensors, and other details that are not explained here. They are useful when looking at the interactive visualization, but are not relevant here. Note that the IN has significantly more parameters than is apparent from this image: Each tensor can have an arbitrary size, and each PU contains a neural network of arbitrary complexity. In fact, in the code I used to generate this, each individual PU internally contains a complete PyTorch Model. This can be inspected in the interactive visualization.}
	\label{fig:paper_1_intro}
\end{figure}

Figure \ref{fig:paper_1_intro} shows a visualization of an Interaction Network.

\subsubsection{Algorithm}

An Interaction Network does not have inputs and outputs in the conventional sense, like normal neural networks do. Instead, the IN is an autonomous agent that runs in an endless loop. Explicitly defined Environments are used to provide inputs and to read outputs at the appropriate time.

The loop of the IN looks like this:

\begin{algorithm}[H]
	\algsetup{linenosize=\tiny}
	\scriptsize
	initialize Nodes
	
	\While{True}{
		\tcc{Receive inputs from the outside}
		\For{$env \in Environments$}{
			\If{env has input from outside}{
				update Nodes with input\;
			}
			set network signals, for the Control Unit to use as input\;
		}
		\tcc{The Control Unit uses a DQN to determine how the IN algorithm proceeds.}
		next\_action $\longleftarrow $Run Control Unit to determine next action\;
		\eIf{type of next\_action is 'Environment'}{
			\tcc{Environments may expose any number of actions. These can be used to interact with the outside world.}
			signal the selected Environment to perform its action\;
		}{}
		\eIf{type of next\_action is 'Processing Unit'}{
			\tcc{When a PU is executed, it takes the current values of its input Nodes as inputs, applies the neural network contained in the PU to that data, and stores the results in its output Nodes}
			execute the selected Processing Unit\;
		}{}
		\tcc{Receive feedback from the outside. This comes in two distinct types: Rewards for the CU, and gradients for the Nodes.}
		\For{$env \in Environments$}{
			\If{env gives reward}{
				\tcc{The CU gets updated in the same way as DQNs, in a non-episodic, infinite-horizon learning process.}
				assign reward to Control Unit\;
			}
			\If{env gives gradient}{
				\tcc{The PU receives a gradient through a traditinal loss function.}
				assign gradient to Node\;
				backpropagate gradient through the sequence of Processing Units that updated the Node\;
			}
		}
	}
	\caption{The loop of an Interaction Network}
\end{algorithm}

The above pseudocode is deliberately vague. This is because there is a lot of flexibility in the implementation details of each component.

For example, the CU basically acts like a DQN in many respects, so all of the many variants of DQNs that have been invented can also be tried here. However, the purpose of the CU (choosing the next algorithm to run) is not identical to the purpose of a DQN (choosing the next action to perform in the environment). Because of this, the optimal model and hyperparameters may be different from those of DQNs. The same is true for other components of the IN.

On each iteration, the CU essentially gets to choose between performing an internal update step (running a PU), or an external interaction with the environment (triggering an Environment).

If the problem the IN is working on is difficult, the IN may take as many internal steps as necessary to solve the problem before submitting a solution.

Unlike a DQN, the CU does not get the state of the environment as input. It takes the following inputs instead:

\begin{itemize}
	\item The history of recently executed actions.
	\item Signals that were set by the Environments.
	\item The current values of Nodes, or of a subset of them.
\end{itemize}

The state of the Interaction Network is defined by the states of its Nodes. These come in different types:

\begin{itemize}
	\item The simplest type of Node is just a slot that holds a single tensor, of arbitrary but fixed size.
	\item More complex Nodes can represent addressable memory systems that connect to an outside data storage. They can store arbitrary amounts of memory. This makes the IN Turing Complete, similar to Differentiable Neural Turing Machines\cite{graves2016hybrid}.
	\item The content of each Node can optionally be available to the CU as input, to help inform its decision making. In the visualizations shown throughout this paper, these Nodes are displayed with two tensors. The first is only available to the PU's, and the second, smaller one, is available to the CU as well. This is done to keep the total size of the CU's input small.
\end{itemize}

The Interaction Network possesses multiple Processing Units. These can perform several different roles, each useful to the IN in different ways:

\begin{itemize}
	\item The results of the PU's can update Nodes, which in turn are used as inputs for the CU. In this way, a CU can execute a PU in order to gain information that then informs its next steps.
	\item A PU can solve a specific machine learning problem, independent of the CU, or other PU's.
	\item A PU can provide intermediate results to enable transfer learning for other PU's.
	\item The results of the PU's can be used to parameterize the actions of Environments.
\end{itemize}

\subsection{The key Interactions}

The Interaction Network is named so because of the way all of its components interact and play off each other:

The CU learns what actions to perform based on a context. The Nodes represent that context. The PU’s can update the context.

The result of this interaction is that the IN learns to associate different circumstances / contexts with different expectations of rewards. In a trained IN, a single execution of a PU can completely shift the reward predictions of the CU and thereby change its behavior.

Since the values of Nodes are persistent and only change if they are updated, these contexts are stable until the CU runs a PU to change it again.

For example: Assume that we have a fully trained Interaction Network whose task is object identification in images. There are many PU’s, and each of them is good at a specific subtask of object recognition. At one point, a PU may recognize that the image has been flipped by 90 degrees. This information is made available to the CU through the output of the PU. The CU now decides to execute a specialized “turn-image” PU to shift the image back into the correct orientation. After this correction, the CU notices that the special case has been resolved, and proceeds with what it was doing before. Because all relevant information is persisted in Nodes, there is no loss of information and the CU can immediately resume where it left off.

This example Interaction Network uses one PU to detect when it is necessary to rotate an image, and a second PU to perform the rotation. As a consequence, all other PU's of the IN can assume that their inputs are always already rotated correctly. In this way, the IN enables rotational invariance, which is a problem that is normally very hard to capture with a CNN\cite{sabour2017dynamic}.

This interaction between Nodes, CU and PU's can be described in an analogy:

The current values of the Nodes represent the mental state of the Interaction Network. The individual Processing Units represent different thoughts. The Control Unit is the mechanism that determines which train of thought has dominance in the mind.

Having a thought (running a PU) can change your mental state (Nodes). Your mental state influences which thoughts become dominant (CU).

\subsubsection{Flexible Architecture}

Interaction Networks have one more unusual property, which is optional:

\emph{Their architectures do not need to be fixed}.

It is not much effort to add or remove Nodes and PU's to an existing IN, even in the middle of training. One just has to ensure that when an object is added that has actions associated with it (e.g. executing a PU), the weights for selecting that action in the CU are initialized at an appropriately low level.

Many usecases of IN's work just fine with fixed architectures. However, some tasks described below rely on Curriculum Learning, and for many of them the learning process may work more effectively if new Nodes and PU's are added later during training, so that they do not distract the CU early on. The goal is to always have a fixed amount of untrained components, and to add new ones whenever the existing ones start being specialized.

\section{Emulating and improving on contemporary neural networks}

To demonstrate what Interaction Networks can look like in practice, I will discuss a number of contemporary types of neural networks. For each of them, I will highlight the following points:

\begin{itemize}
	\item \emph{Emulation}: How could an Interaction Network emulate the behavior of the neural network, assuming that it is already fully trained?
	\item \emph{Challenges}: How can you train an IN to reach that point? Will the IN be able to learn as efficiently as the algorithm it is emulating? What additional problems start to appear when you stop trying to emulate the NN strictly and try to make use of the IN's flexibility?
	\item \emph{Advantages}: How can the increased flexibility of an IN be used to improve the emulated algorithm further? Can this help to mitigate or overcome any of the challenges?
\end{itemize}

\subsection{FNN: Feedforward Neural Network}

Let us consider the most basic type of neural network as a first example. This is just a shallow network that maps one input vector to one output vector.

\subsubsection{Emulation}

\begin{figure}[H]
	\centering
	\fbox{
		\includegraphics[width=0.5\columnwidth]{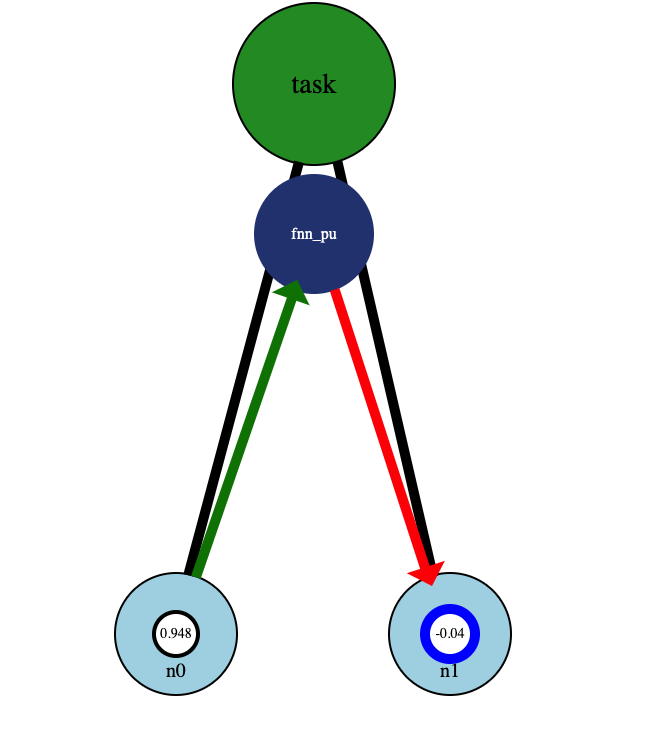}
	}
	\captionsetup{singlelinecheck=off}
	\caption[list=off]{The Environment 'task' delivers the input to Node n0. The PU fnn\_pu contains the FNN and applies it to n0. It writes the result into n1. The Environment then reads n1 and applies a gradient to it.}
	\label{fig:paper_1_fnn}
\end{figure}

Figure \ref{fig:paper_1_fnn} illustrates how to emulate an arbitrary FNN in an Interaction Network.

\begin{figure}[H]
	\centering
	\includegraphics[width=0.5\columnwidth]{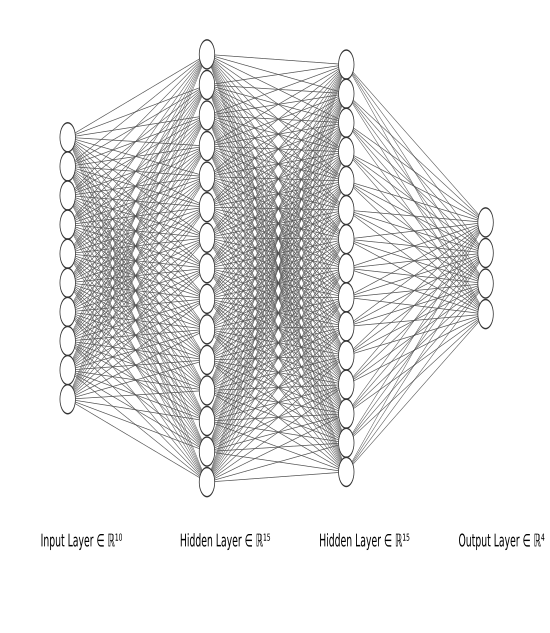}
	\caption{This entire neural network is contained in the PU called "fnn\_pu" of Figure \ref{fig:paper_1_fnn}. The input layer receives the Node n0, and the output layer is written into the Node n1. This is an example to demonstrate that each PU contains many parameters. PU's can contain arbitrary neural networks, and therefore can be arbitrarily complex.}
	\label{fig:standard}
\end{figure}

\subsubsection{Challenges}

The above example emulates the FNN perfectly. Both the feedforward execution and the training through backpropagation work exactly as they normally do.

Challenges only appear if we try to improve things:

In a proper Interaction Network, there should be several PU's and several Nodes, to enable interactions between different neural networks.

If multiple Nodes and PU's are present in a network, then things become more complicated:

\begin{itemize}
	\item If there is no PU that directly connects the Environment's input Node with its output Node, then the CU has to perform a sequence of PU's in the correct order to create a path from the input Node to the output Node. Executing PU's in an incorrect order will waste time. It can also make the results worse, and can even slow down the training process because the backpropagation algorithm will be applied to the wrongfully executed PU as well, which trains it in the wrong direction.
	\item Learning the correct order of execution of PU's is difficult for two reasons:
	\begin{itemize}
		\item The CU only learns that a sequence of PU executions was helpful after executing all of them. This is the usual problem of sparse and distant feedback in Reinforcement Learning.
		\item While the PU's are untrained, executing the correct sequence of PU's will not actually lead to a high reward. Therefore, the IN has to train both the CU and all the PU's in parallel. This may take a prohibitively long time to converge.
	\end{itemize}
	\item Backpropagation of losses to the PU's can become computationally expensive if a long sequence of PU's was involved in setting the value of a Node. Worse, because a Node may at times be set by different sequences of PU's, the gradients received by each PU can be highly variable. In addition to the usual problems of Exploding Gradients and Vanishing Gradients, one may encounter Variable Gradients. These confuse most commonly used optimizers, which assume that the gradients they operate on are always produced in the same way, and therefore scaled in the same way. Normalizing gradients will play an even more important role in Interaction Networks than in normal Neural Networks.
\end{itemize}

\subsubsection{Advantages}

All of the above challenges are caused by the increased flexibility of Interaction Networks compared to FNNs. If these challenges can be overcome, then this flexibility will offer a lot of advantages:

\begin{itemize}
	\item The IN can learn to train multiple systems of PU's in parallel, each trying to solve the same task in different ways. The CU submits whichever one of their results seems best, or interpolates them through yet another PU. A properly tuned IN should therefore be able to emulate Ensemble Learning all on its own, without explicitly coding for it. I see this as a likely low-hanging fruit to achieve good results with an IN.
	\item An IN's architecture is uniquely suited to take advantage of Curriculum Learning. If the IN is taught gradually more complex tasks, then the IN can train one new PU at a time, while relying on the intermediate results of already trained older PU's. This is a form of Transfer Learning. Such a curriculum should also make it much easier for the CU to learn a sequence of PU's, since the sequence can be learned piecewise instead of all at once. Unlike normal neural networks, it is theoretically even possible to let an IN choose its own curriculum by letting it request training tasks on its own, through Environments. This would require rewarding the CU based on uncertainty reduction, which is a promising avenue of further research for IN's.
	\item The harmful interference between PU's, wherein PU's share output Nodes and overwrite each others' results, can be turned into a positive by carefully tuning the way in which PU's are created and the way they connect Nodes. Once properly configured, the previously destructive overwriting of Nodes turns into a beneficial corrective process: More specialized PU's can learn to veto more general PU's by overwriting their results when necessary. This may be made easier by giving Nodes a memory, so that they can store and interpolate between multiple recent values instead of only keeping the last value that was written into them.
	\item Deeper networks can be split into smaller sections by turning hidden layers into Nodes. If the features learned by these hidden layers are useful, then explicitly representing them as Nodes makes those features accessible to other neural networks. This is a form of Transfer Learning. The drawback is that this brings us back to one of the challenges: For each PU that is split into smaller parts, the CU now has to learn a longer sequence of actions, because it needs to learn that one of these parts should be executed before the other.
	\item By constructing parallel PUs that compete with each other, the IN effectively performs neural architecture search as a byproduct of its learning process. It is possible to generate entire new PU’s at runtime in order to explore alternatives. The challenge is finding a good heuristic for determining the tradeoff between exploration (creating and training new alternative PU’s) and exploitation (fixing PUs in place that have already proven to be useful.)
\end{itemize}

\subsection{CNN: Convolutional Neural Networks}

Convolutional Neural Networks are the go-to approach for image processing tasks of all kinds.

\subsubsection{Emulation}

An IN can emulate a CNN in the same way as an FNN. A Processing Unit can contain Convolutional Layers or Max-Pooling layers without issue.

\subsubsection{Challenges}

The challenges are the same as for an FNN.

\subsubsection{Advantages}

The flexibility of IN's could prove to be very useful for object detection. One of the major challenges in object detection is that it is essentially composed of two separate tasks: Identifying candidate regions of the image, and scanning those regions for objects.

Previous work has combined both of these tasks into a single step, with great success\cite{girshick2015fast}.

However, it is curious that the object detection of humans and animals appears to work differently: We do not notice objects in our field of vision all at the same time.

The longer we stare at a picture, the more details we become aware of. Identifying camouflaged animals, or seeing through optical illusions, can take several seconds of analysis. In this time, we notice one detail after another, and each detail makes us more able to recognize the next\texttt{rest}. Once you have identified part of an image as the tail of an animal, your attention is automatically drawn to where you suspect the rest of the animal's body to be, and it becomes easier to notice the rest of the animal.

An Interaction Network should be able to replicate this behavior: It can look at the same image multiple times, focusing on different things each time, and every piece of information it obtains can help guide it to notice other things.

\begin{figure}
	\centering
	\fbox{
		\includegraphics[width=0.8\columnwidth]{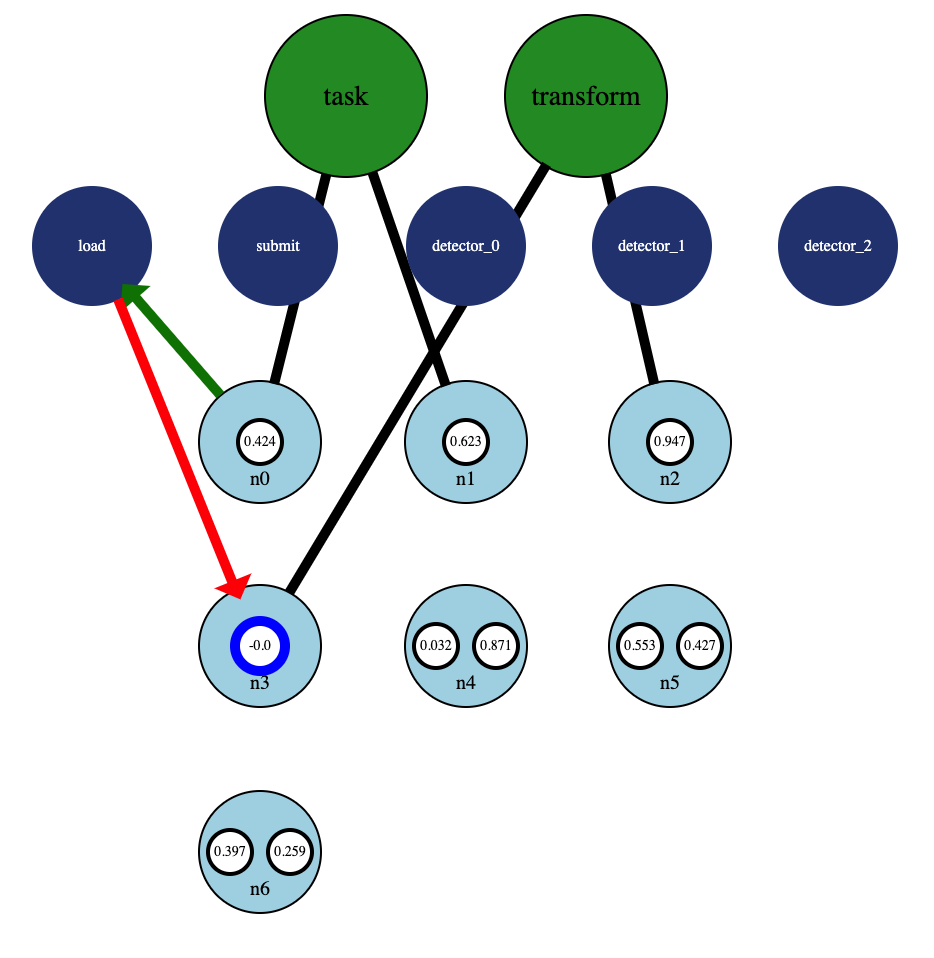}
	}
	\captionsetup{singlelinecheck=off}
	\caption[list=off]{n0 is where data is provided, and n1 where the result is submitted. n3 is used as the primary Node to work on. Some connections of Processing Units were omitted in this image to make it easier to read:
		\begin{itemize}
			\item 'submit' reads from n3 and writes to n1.
			\item The 'detector' PU's each read from n3 and write to n2.
			\item Each of the detectors also writes to one of n4, n5, or n6. These Nodes come to represent features that are detected, and which influence the Control Unit.
		\end{itemize}
	}
	\label{fig:paper_1_modular_cnn}
\end{figure}

Figure \ref{fig:paper_1_modular_cnn} illustrates how this would work.

\begin{itemize}
	\item The Interaction Network has access to an Environment that can zoom into an image. This Environment has one Node that the CU can use to specify a region to zoom in on, and one Node in which the Environment returns the zoomed image.
	\item There exist multiple PU's for detecting objects. Some of these are generic object detectors that operate on the entire image. Others are specialized units that try to detect objects given that certain assumptions are met. For example, one PU could be specialized in detecting cars and traffic signs, and it is only used if a more generic PU indicates that the image's overall color scheme matches that of a city.
	\item The CU tackles each new object detection task by running a generic detector first. Depending on the results of this generic detector, it may then decide to run more specialized PU's, or it may use the Environment to zoom into particular regions of interest to focus on them. This continues until the IN is confident that it has detected all there is to find, and submits its results. By applying a penalty to false positives or to the number of iterations spent searching, the network can be tuned to spend only as much time as necessary on each image.
\end{itemize}

In this way, an Interaction Network could learn to generate PU's with the specific purpose of enabling invariances in object detection. Some invariances in images are normally very hard to model with CNN's, which often have difficulty adjusting to novel view points\cite{sabour2017dynamic}. An Interaction Network, in contrast, can learn separate PU's for detecting image perturbations, fixing them, and detecting objects in the fixed images. This is much more effective than learning an exponentially growing number of features for every possible type of image perturbation.

This system will likely not be very fast to train since it contains so many interacting components. However, its performance during detection should be both fast and reliable, especially since you can explicitly give the CU a time constraint to tell it how much time it has to explore different possibilities. The CU may be taught to only investigate the most promising things, or to take its time and investigate every possibility.

I expect that the major difficulty in training such a system will come from the large number of interactions between different components. Fortunately, Curriculum Learning should work perfectly for this problem:

Start with a simple IN that just replicates a normal CNN. Train that model on very simple, centered images. Then add the Environment and the zooming functionality, and show it images that have been altered. Then provide gradually more complex special cases, and let the CU develop detection mechanisms and specialized PU's to deal with them.

Each of these steps is relatively simple on its own. So long as you train long enough for earlier steps to converge, a well-chosen curriculum should allow an Interaction Network to learn to detect even very complex objects.

Note that this procedure relies on using an adaptive architecture for the IN, instead of a fixed one. New Nodes and PU's must be added over time. A good heuristic to use would be to keep track of the number of Node's and PU's that are already actively involved in solving problems compared to those that are not yet trained. Whenever the ratio rises, add more Nodes and PU's to the IN. That way, the IN always has a limited number of untrained PU's and the CU doesn't get confused by too large a number of actions that have not yet become established.

\subsection{RNN: Recurrent Neural Networks}

RNN's are useful whenever we want to operate on data that is best expressed as a sequence of items.

\subsubsection{Emulation}

\begin{figure}
	\centering
	\fbox{
		\includegraphics[width=0.5\columnwidth]{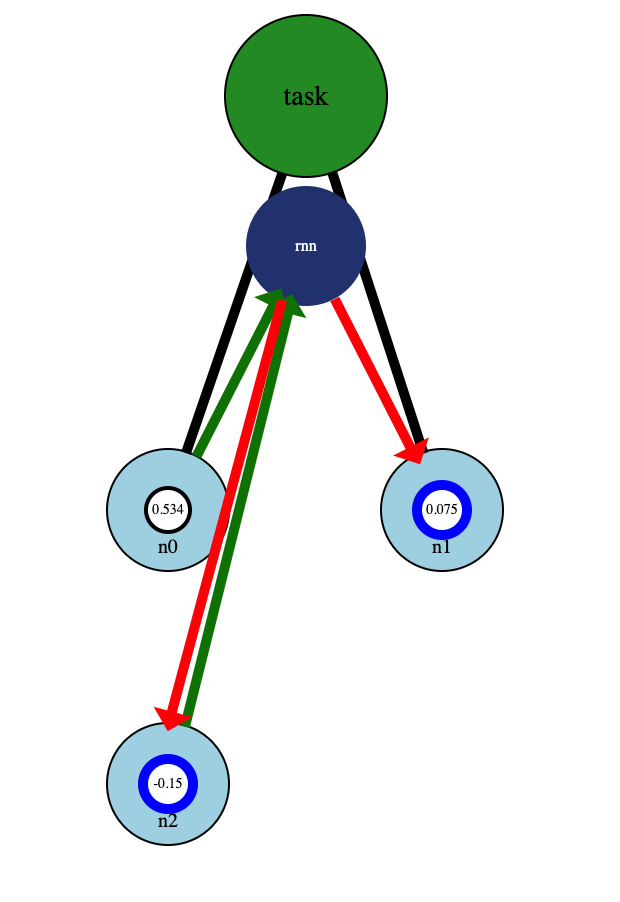}
	}
	\captionsetup{singlelinecheck=off}
	\caption[list=off]{
		The Environment 'task' delivers the input to Node n0. The PU rnn contains the RNN and applies it to n0. It writes the result into n1. The Environment then reads n1 and applies a gradient to it. The Node n2 acts as the RNN's memory: It is used as both input and output each time.}
	\label{fig:paper_1_rnn}
\end{figure}

Figure \ref{fig:paper_1_rnn} shows an emulation of an RNN.

\subsubsection{Challenges}

RNN's are a natural fit for Interaction Networks. The memory of the RNN can simply be modeled as a Node of the IN.

We just need to be careful to ensure that the CU learns one simple rule: It should execute the PU exactly once per new data point that arrives through the Environment. Otherwise the emulated RNN will either skip inputs, or read the same input multiple time.

There are also some minor issues besides that, like preventing endless loops or resetting the memory when a new sequence starts, but these problems are comparatively simple to solve.

\subsubsection{Advantages}

Making the memory of the RNN a Node comes with a number of advantages. Other PU's can connect to it, and make use of it. It is possible to create schemes of many PU's that share different Nodes with each other. This allows for the creation of complex memory systems that can operate at different levels of time resolution, can be context-sensitive, and can even include associative memory blocks.

It should be obvious that such a complex memory system would be superior to an RNN once it is fully trained. The equally valid counterpoint is that such a system would be too hard to train to be useful in practice.

However, while such an extreme system would likely be too complex to work, I find it highly likely that there is some less extreme modification, using only a subset of these features, that would lead to performance gains in practice.

In fact, the next algorithm to discuss, Attention Mechanisms, \emph{are} in fact using a subset of these components, and they have been shown to give better results than RNN's on many tasks.

\subsection{Attention Mechanisms}

Attention Mechanisms are a complex type of neural network that captures the idea of attention, i.e. of taking more note of some concepts than of others. This improves substantially on RNN's in the domain of machine translation, and has shown great promise in other domains as well.

\subsubsection{Emulation}

\begin{figure}
	\centering
	\fbox{
		\includegraphics[width=0.5\columnwidth]{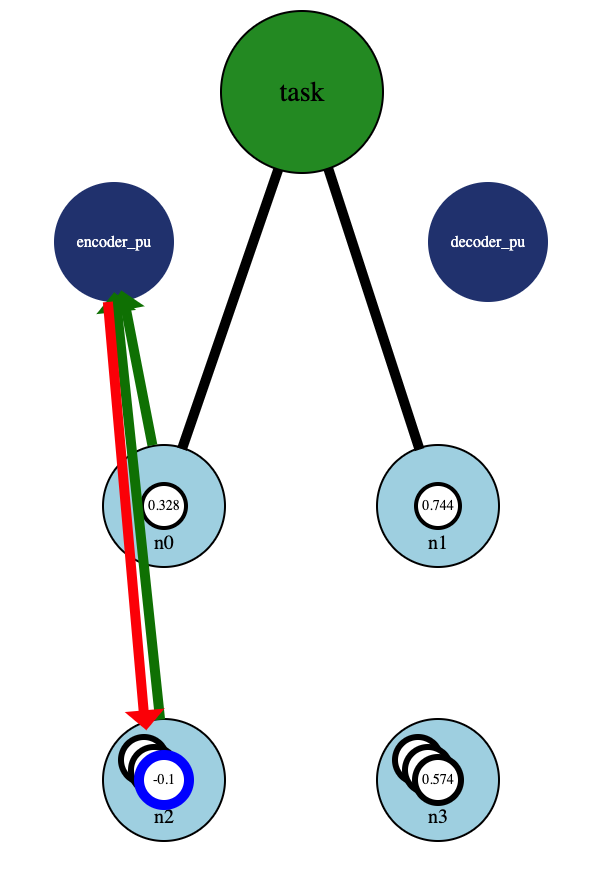}
	}
	\captionsetup{singlelinecheck=off}
	\caption[list=off]{The Environment uses n0 to present the next input word, and n1 to retrieve the next output word. The Nodes n2 and n3 hold not just one tensor like before, but an arbitrary number of tensors. The decoder has n2 and n3 as inputs, and n1 and n3 as outputs (not visualized here).}
	\label{fig:paper_1_attention}
\end{figure}

Figure \ref{fig:paper_1_attention} demonstrates how to emulate an Attention Mechanism in an Interaction Network. This illustration hides most of the complexity of Attention mechanisms. The details are not important here, just the main layout.

The overall approach is mostly the same as for RNN's, but it is much more complex because an Attention Mechanism has a lot more steps that need to be replicated in the right order.

This network makes use of a more complex type of Node than previously discussed. To replicate an Attention Mechanism, it is necessary to store the encodings of all incoming word vectors at the same time, since some operations depend on combining all of them. To do this, we introduce a new type of Node that can accumulate multiple entries, and pass them to a PU as a batch. The CU gets an action to clear the accumulated entries of the Node.

\subsubsection{Challenges}

It would be very difficult to train an IN to emulate an Attention Mechanism. The actions need to be performed in just the right order, and the mapping of the encoder and decoder states in the PU's needs to be just right.

Attention Mechanisms are likely too specialized to arise naturally in an IN through training.

However, they are useful for illustrative purposes, because there are similar architectures that an IN, \emph{could} learn on its own. Even though this particular architecture is too complex to expect an IN to learn it, the fact that it is possible to emulate it at all suggests that there may be low-hanging fruit out there: Architectures that are somewhere between an RNN and an Attention Mechanism in complexity, that perform well in practice, and that an IN could find on its own.

\subsection{DQN: Deep Q-learning Network, and Policy Gradient Methods}

Deep Q-learning Networks and Policy Gradient Methods are two popular methods of Reinforcement Learning based on similar approaches, but with different underlying principles.

\subsubsection{Emulation}

Emulating DQN's is a special case of Interaction Networks. Since the Control Unit \emph{is} a DQN, it is actually possible to emulate a DQN by building an Interaction Network without any Nodes or PU's.

Of course, doing this would defeat the entire point of using an Interaction Network at all.

\begin{figure}
	\centering
	\fbox{
		\includegraphics[width=0.5\columnwidth]{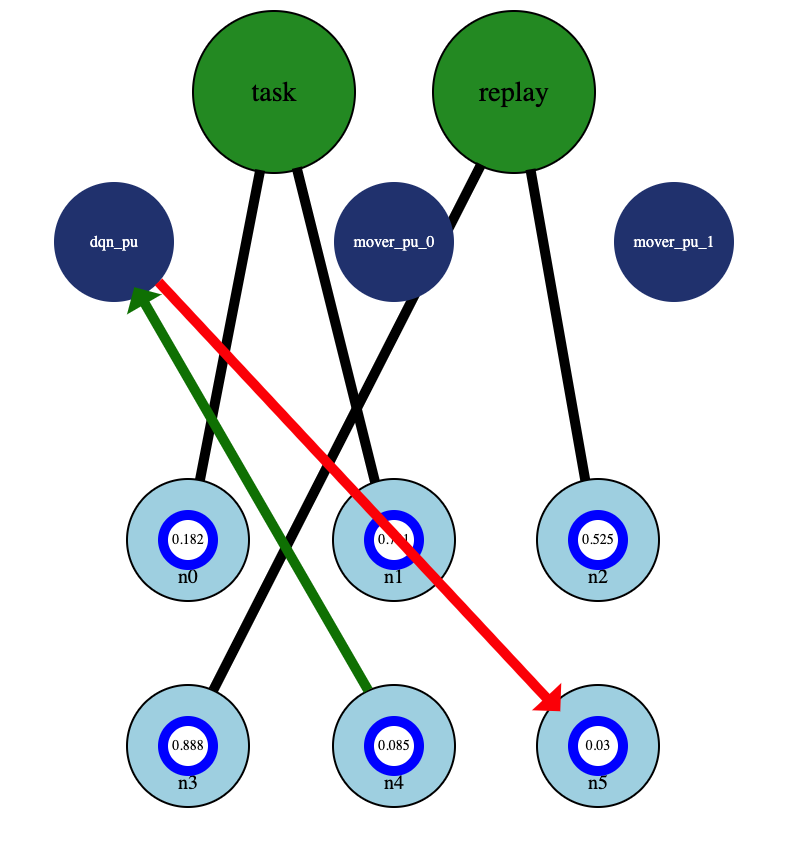}
	}
	\captionsetup{singlelinecheck=off}
	\caption[list=off]{n0 is where data is provided by the main environment, and n1 where the result is submitted. n2 and n3 serve the same role for experience replay. n4 and n5 serve as the input and output Nodes for both of these. The 'mover' PU's don't perform any processing, and just move inputs from n0/n2 to n4 and from n5 to n1/n3. They are essentially just helper functions that the CU can use to move data around before processing it.
	}
	\label{fig:paper_1_dqn}
\end{figure}

Instead, it is also possible to implement a DQN as a PU within the IN, as illustrated in figure \ref{fig:paper_1_dqn}.

We end up with two DQN's: The one in the PU, which decides which actions to perform in the game environment, and the one in the CU, which just executes the PU when new data arrives, and decides whether to use the Environment for Experience Replay, or the real one.

The reward signals become rather unusual in this setup: The PU gets a reward based on how useful its actions in the environment were. The CU gets a reward based on how good of an idea it was to let the PU pick the action, and on which Environment to work. Since the PU is the only action available to the CU, the rewards are determined only by the choice of Environment: Experience Replay or live play.

Notably, both the CU and the PU can be optimized independently: Using Dueling Double DQN, adjusting the decay factors, and other improvements can all be applied to each of them separately.

The exception to this is Experience Replay: The PU can not have Experience Replay, on its own, since the CU dictates its inputs. This is why a second Environment needs to be attached to the IN, which is in charge of storing and loading experiences. The CU actually has its own analogue for Experience Replay called Scenario Mechanisms, but these are an advanced topic that will be introduced in a later section of this paper.

It is also possible to use a Policy Gradient Method in the PU instead of a DQN. The IN does not need to be changed for this, assuming we use a variant of Policy Gradient that can learn off-policy, so that using Experience Replay is sensible.

\subsubsection{Challenges}

The challenges are mostly the same as for an FNN. As far as the Interaction Network is concerned, the DQN can be treated like any other FNN.

You can tune the reward given by the Experience Replay Environment to the CU in order to control how often the CU performs experience replay, and how often it tries to solve live problems.

If the CU reward of the Experience Replay Environment is based on reducing a measure of uncertainty, this would actually allow the IN to learn to use Experience Replay only when there is an experience that is worth replaying, which could increase the learning speed (in practice the overhead of running the CU would probably outweigh any gains here, though).

\subsubsection{Advantages}

The advantages for FNN's also apply to DQN's.

There is thing that is worth pointing our specifically however:

A well-tuned IN using Curriculum Learning is likely to develop an analogue of Hierarchical Reinforcement Learning on its own.

This is similar to the process outlined for object detection with CNN's: The Interaction Network will first learn a simple DQN. It will then learn to detect special cases. Based on those special cases, it can now construct more specialized PU's that support and enrich the generic DQN's if the CU decides that they are appropriate for the situation.

\section{Miscellaneous observations}

Besides reproducing different types of neural networks, Interaction Networks have interesting relationships with various other methods and concepts from machine learning and artificial intelligence in general.

The following lists some of these relationships that may be interesting to consider:

\begin{itemize}
	\item Multi-task Learning: Combining multiple types of data in a single model has been shown to give surprisingly good results\cite{kaiser2017one}. The modular nature of Interaction Networks should be well-suited to this.
	\item Ensemble Learning can be emulated by first training multiple PU's on the same objective independently, then training one final PU to combine their results. Theoretically, a CU could be trained to understand this order of events in general. If this is successful, an Interaction Network could be trained to understand the process of building an Ensemble, and would then be able to autonomously use Ensemble Learning in the future. This is pure conjecture at this stage and likely very difficult, but may be worth exploring.
	\item Neural Architecture Search: If the Nodes and PU's of an Interaction Network are not statically defined but dynamically changeable, the IN performs neural architecture search implicitly in the course of its normal operations. The mechanisms to achieve this effectively are beyond the scope of this paper and a topic of future work.
	\item Interaction Networks are a hybrid of Connectionist and Symbolic AI. Neural networks are connectionist. Nodes and external memory structures can store symbolic data. The fact that PU's are discrete objects likewise is a form of symbolic information. For example, the pattern “PU1, then PU2, then either PU3 or PU4 depending on condition X” is symbolic, but can be learned through Connectionist learning processes in the CU. By encoding values in a node that control what PU runs next, you can simulate hierarchical knowledge storage, which is very important for symbolic reasoning.
	\item Memory-augmented Neural Networks (MANN's): Adding memory blocks to neural networks has been shown to improve performance on some tasks. There are different types of memory-augmented neural networks, with different ways of storing an referencing memory blocks. Each of these can be emulated by an Interaction Network by using Nodes to act as memory blocks. The big issue here is differentiability. Existing MANN's take great care to ensure that their memories are end-to-end differentiable. Ensuring this in the general case of Interaction Networks may get tricky. Because of the unpredictable order of input-function-output sequences of PU's, backpropagating through an IN requires tracking the gradients explicitly, even though memories may in principle be stored for arbitrary amounts of time.
\end{itemize}

\section{First Experiments}

I have performed a number of simple initial experiments to test how well Interaction Networks work on toy examples.

These experiments were not intended to solve a useful problem. Rather, the purpose was to compare how the learning behavior of the Interaction Network changes based on the way the network is configured, and the problem is presented. These examples are simple, and assume that the IN has a static and predefined set of Nodes and PU’s.

Note that all of this uses vanilla DQN, and standard values for hyperparameters. Using the myriad improvements to DQN used in SOTA systems would likely improve results substantially. However, the purpose of this section is to analyze the strengths and weaknesses of Interaction Networks on their own, so I thought it best to keep everything else as simple as possible for now, even if this lowers performance.

I tried two tasks: The first was to learn two separate functions in parallel. The second was to learn a sequence of functions.

\subsection{Experiment 1: Multi-task learning and choosing between alternatives}

For this experiment, we provide the IN with two input values. Both are tensors of size 1. We expect only one output value. If the first input is larger than the second, then the result should be twice the first input. Otherwise the result should be half the second input. The reward given for the task depends only on the mean squared error between the target value and the received value, and ignores which PU delivered the result.

The following approach solves this problem perfectly:
\begin{itemize}
	\item Execute a PU that reads both inputs and compares them.
	\item Choose one of two PU's to execute based on which of the inputs is greater. The first of these doubles the first input and submits it as the solution. The second halves the second input and submits it as the solution.
\end{itemize}

\begin{figure}[H]
	\centering
	\fbox{
		\includegraphics[width=0.5\columnwidth]{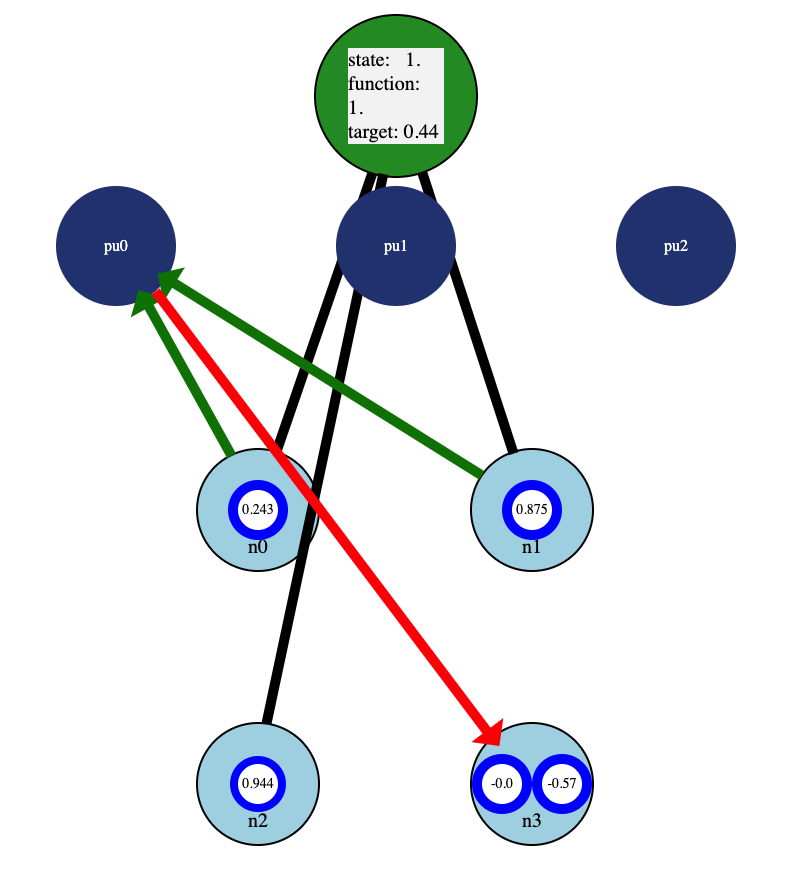}
	}
	\captionsetup{singlelinecheck=off}
	\caption[list=off]{n0 and n1 are used for receiving the two inputs for the task. n2 is used to submit the result. Connections that are not shown: pu1 maps n0 to n2. pu2 maps n1 to n2. The text on the Environment is part of the interactive visualization and helps with debugging, by showing the current status of the task.}
	\label{fig:paper_1_cra}
\end{figure}

The Nodes and PU's are created in such a way that their layout makes this easy to learn, as shown in figure \ref{fig:paper_1_cra}. However, neither the CU nor the PU's are correctly trained in the beginning.

\subsubsection{Variant 1: Base case}

Despite how simple this task is, training it is deceptively complicated.

The following things need to be learned before optimal behavior is possible:
\begin{itemize}
	\item The CU has to react to new input by calling pu0.
	\item pu0 has to compare which of its inputs is larger.
	\item The CU has to look at the results of pu0 and choose which of pu1 and pu2 it should call next.
	\item Both pu1 and pu2 need to be trained correctly.
\end{itemize}

Until all of these conditions are true, it is easy to fall into a local optimum.

Worse: Local optima can be difficult to unlearn. pu1 and pu2 receive a gradient whenever they are used to submit a result. This is true even if they are used incorrectly. This can cause them to be trained wrong. If they are trained wrong, then this in turn means that the theoretical optimal behavior does not receive the optimal reward. This in turn means that pu0 and the CU do not receive the right kind of feedback, either.

\paragraph{Results}

\begin{figure}[H]
	\centering
	\fbox{
		\includegraphics[width=0.5\columnwidth]{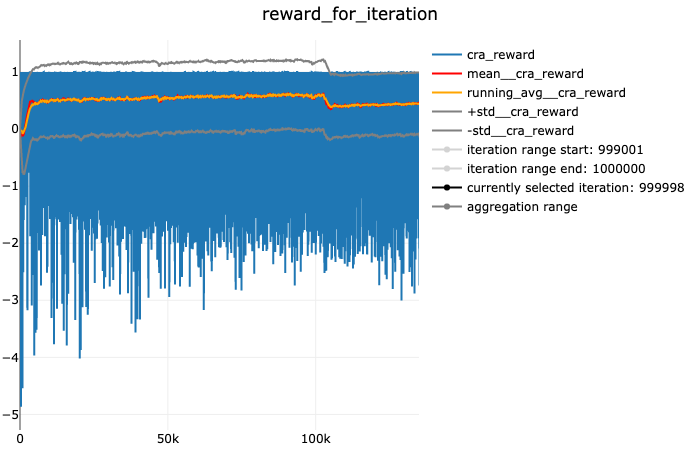}
	}
	\captionsetup{singlelinecheck=off}
	\caption[list=off]{This image shows the average reward the Environment gave, over time. The relevant part is the orange line. Note that this image is somewhat unintuitive to interpret: It appears that the performance increases very quickly to an average of 0.5, then keeps rising. At around iteration 100,000 it suddenly drops, before rising again to 0.5, where it remains stable.
		
	What happened is this: The IN quickly learned to just always use pu1 for every problem. This is correct in 50\% of the cases. In the remaining 50\% of cases where pu2 should have been used instead, the result is still sometimes not entirely wrong, so the average reward is greater than 0.5. Since it only requires one step of the IN to do this, the IN receives an average reward of greater than 0.5 per iteration. This is a local optimum.
	
	Around iteration 100,000 the IN broke out of this local optimum as a result of random exploration (the CU sometimes takes random actions instead of the highest rated ones). It then converged to the actual optimal behavior: Run pu0 first, and then either pu1 or pu2. However, since this requires two steps of the IN, the graph makes it look like the performance is worse, since the average of 0 (no reward on the iteration where pu0 is executed) and 1 (perfect reward on the second iteration) is 0.5.
	
	I left this graph as it is on purpose, because it is just one of several counterintuitive findings I encountered while training IN's. The fact that IN's can freely choose to ignore the task for many iterations makes them difficult to debug, and makes many usual KPI's unreliable. This is why the interactive visualization from which this graph was taken is necessary to understand the IN's behavior more easily. Unfortunately, that visualization can't be attached to a PDF or printed on paper.}
	\label{fig:paper_1_cra_1_rewards}
\end{figure}

Figure \ref{fig:paper_1_cra_1_rewards} shows that the task is learned correctly. However, the IN gets stuck in a local optimum for a full 100,000 iterations, before it learns the correct solution.

Over 100,000 iterations seems excessive for learning to solve such a simple problem, even if most of that time was spent stuck in a local optimum. Let's try to make this experiment easier, and see if that speeds up the learning process.

\subsubsection{Variant 2: Inputs not available to CU}

For this experiment, we make the two input values available to the CU directly. This means that it is now possible to just skip the execution of pu0, since the CU can compare the two inputs internally.

Removing a step from the optimal sequence of steps like this should make it much easier to learn.

\paragraph{Results}

\begin{figure}[H]
	\centering
	\fbox{
		\includegraphics[width=0.5\columnwidth]{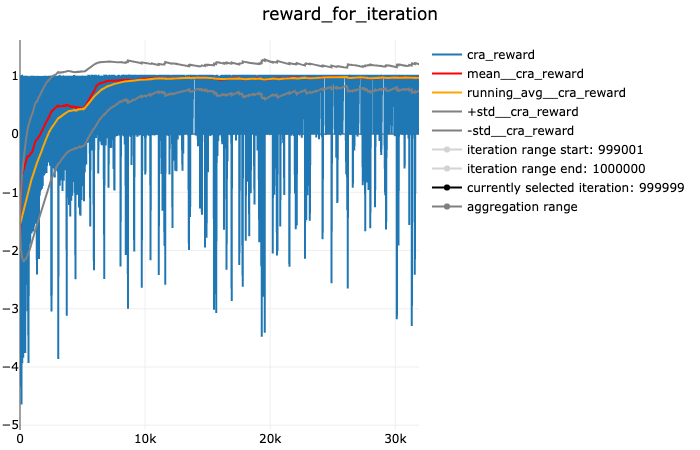}
	}
	\captionsetup{singlelinecheck=off}
	\caption[list=off]{The IN converges to a perfect reward much more quickly. Note that a perfect result is now a reward of 1, not a reward of 0.5, since it now only takes one iteration to succeed at the task.}
	\label{fig:paper_1_cra_2_rewards}
\end{figure}

This experiment showed that the IN now learns the task very quickly and converges to the optimal strategy in only about 10,000 iterations. Curiously, we still briefly get stuck in a local optimum around 5000 iterations. See Figure \ref{fig:paper_1_cra_2_rewards}.

\subsubsection{Variant 3: Training wheels for the CU}

Since it seems that learning both the execution order and the PU's own networks at the same time is difficult, let's see if training gets easier if we eliminate one of these problems.

For this experiment, I gave training wheels to the CU: For the first n iterations, the CU ignores the reward of the Environment. Instead, I define a hardcoded sequence of actions, which would be the optimal sequence if the PU's were already fully trained, and reward the CU for following this sequence. In this way, the training of the CU is decoupled from the training of the PU's.

\paragraph{Results}

\begin{figure}[H]
	\centering
	\fbox{
		\includegraphics[width=0.5\columnwidth]{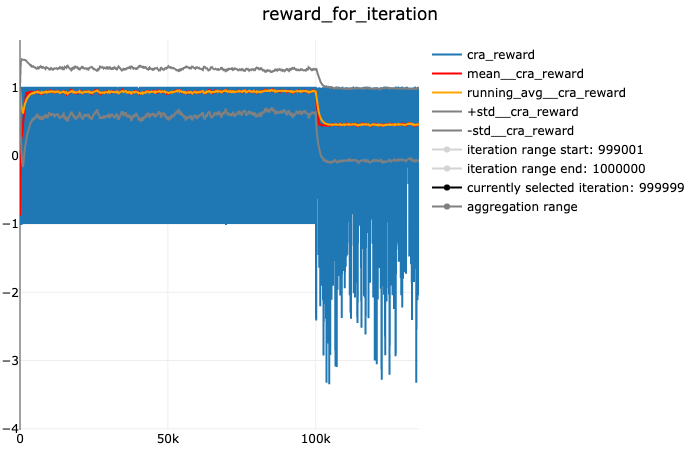}
	}
	\captionsetup{singlelinecheck=off}
	\caption[list=off]{The training wheels cause the IN to learn the problem very quickly (5000 iterations). Performance remains at the optimum once the training wheels come off. Since the fake-reward given by the training wheels was always 1, but the average reward afterwards is 0.5, the CU is briefly disrupted and needs to recalibrate when the training wheels come off at iteration 100,000. This did not cause a problem here, but I imagine it could lead to unexpected results if one is not careful with this while working on more complex tasks.}
	\label{fig:paper_1_cra_3_rewards}
\end{figure}

See Figure \ref{fig:paper_1_cra_3_rewards}.

\subsubsection{Variant 4: Pre-trained PU's}

For this experiment, we address the same problem as in the last experiment in a different way: Training the CU and the PU's simultaneously is difficult. This time we pre-train the PU's to their perfect state before starting the training.

pu0 is trained to differentiate which of the two inputs is larger. pu1 and pu2 are trained to double and half their inputs, respectively.

\paragraph{Results}

\begin{figure}[H]
	\centering
	\fbox{
		\includegraphics[width=0.5\columnwidth]{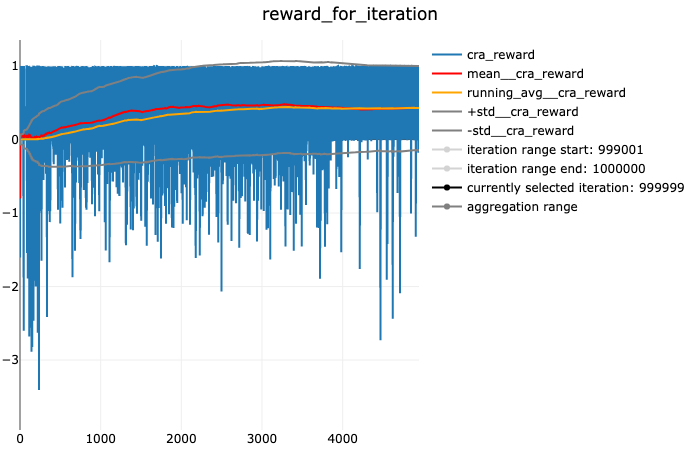}
	}
	\captionsetup{singlelinecheck=off}
	\caption[list=off]{With pre-trained PU's, the Interaction Network converges very quickly to its optimal policy. The CU just has to learn to call pu0 first, and then call pu1 or pu2 based on what it says.}
	\label{fig:paper_1_cra_4_rewards}
\end{figure}

See Figure \ref{fig:paper_1_cra_4_rewards}.

Pre-training the PU’s essentially reduces the problem of training the IN to training a conventional reinforcement learner, with one caveat: Every incorrect action of the CU will cause incorrect feedback to be propagated to the PU’s. In extreme cases, if the CU takes too long to learn, then the PU’s may get destabilized before the CU converges, which causes the entire network to diverge. We will encounter this problem later, in experiment 2.

\subsubsection{Summary}

Interaction Networks can successfully learn tasks where one of several functions must be chosen.

It is very hard to optimize both the CU and the PU’s at the same time, and the Interaction Network is likely to fall into local optima in the process. However, if either the CU or the PU's are already defined, then the IN can learn the other much more easily.

Pre-trained PU's tend to help more than a pre-trained CU. If PU’s are fixed and only the CU learns, the learning process is equivalent to reinforcement learning. If the CU is fixed and PU’s learn, the learning process is equivalent to backpropagation through a DAG of multiple neural networks.

This means in practical terms that transfer learning is likely going to be very important for Interaction Networks, since transfer learning effectively means training one PU at a time.

I did not test this for more complex functions yet, but I see no reason why more complex PU's should substantially influence the convergence speed. Most of the slowdown does not come from the complexity of the PU's but from the fact that training CU and PU's in parallel often leads to incorrect feedback.

\subsection{Experiment 2: Repeated function application}

The experiments so far only required the CU to learn under which circumstances to use which PU, but solving a task never required more than two actions. In this experiment, we want to test if an Interaction Network can also learn longer sequences of actions.

The task is to receive a list of items one by one, and repeatedly apply a function to both it and a memory Node until the sequence is done, then submit the result. The function that is applied must be learned.

Essentially, the CU must learn to simulate "reduce(reductor\_function, input\_list)", while the PU must learn reductor\_function.

The following approach of the CU solves this problem perfectly:
\begin{itemize}
	\item Reset the internal state (a Node)
	\item Repeat until the Environment signals the end of the sequence:
	\subitem Get the next value of the sequence. This is implemented as a CU action that asks the Environment for another input, rather than executing a PU.
	\subitem Execute the PU to perform the function calculations.
	\item Submit the result by putting it into the output Node.
\end{itemize}

\begin{figure}[H]
	\centering
	\fbox{
		\includegraphics[width=0.5\columnwidth]{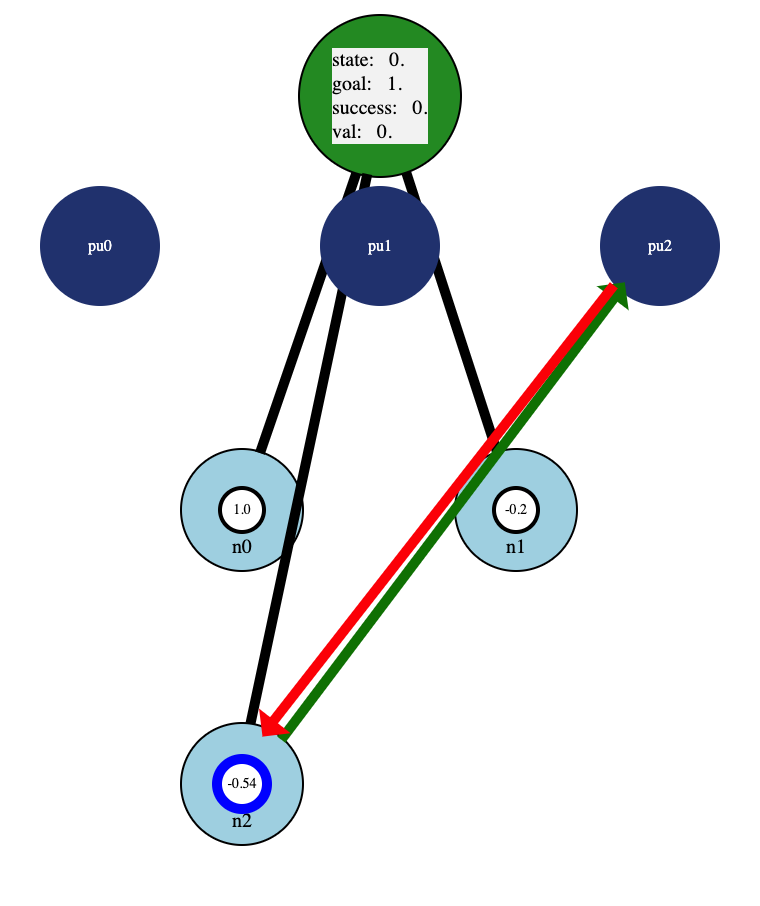}
	}
	\captionsetup{singlelinecheck=off}
	\caption[list=off]{n0 is for receiving the sequence values, n1 is for delivering the final output, and n2 is for performing intermediate calculations. pu0 connects n2 to n1 and is used for submitting the result. pu1 connects n0 and n2 to n2 and is used for learning the function that should be applied repeatedly. pu2 connects n2 to n2 and should be used to reset the internal state at the start of a new sequence.}
	\label{fig:paper_1_rfa}
\end{figure}

The Nodes and PU's are created in such a way that their layout makes this easy to learn, as shown in figure \ref{fig:paper_1_rfa}. However, neither the CU nor the PU's are correctly trained in the beginning.

The function to be learned can be simple or complex. In the simplest case, it's just a constant, so that the input can be ignored. In the most complex case I tried, it is the XOR function. If it is the XOR function, then this task is equivalent to calculating the parity of the input sequence.

\subsubsection{Training procedure}

To train the Interaction Network on this task, we ask the IN to submit a solution to a sequence of length n.

If the IN performs an invalid action, such as requesting another input when the sequence is already finished, it is counted as a failure. Likewise, if the output is incorrect, it is counted as a failure (the output is a binary variable, and is rounded to the closest of the two possible values).

If the IN solves the task at a sequence length n several times in succession, we increase the required sequence length.

In the final evaluation of this task, we are not interested in the reward the IN receives, but in the maximum sequence length for which the IN can reliably solve the task.

\subsubsection{Results}

As it turns out, this Experiment is too complex for the Interaction Network to learn properly without extensive tuning, even when the function to learn is very simple.

However, detailed exploration of many different variations of this Experiment revealed a number of useful facts.

Firstly, the training process sometimes goes wrong in interesting ways that do not happen in normal neural networks.

Secondly, many of these problems should be solvable through a small number of improvements, which will be summarized later in this paper.

The following are the most interesting findings:

\paragraph{Reward tuning}

Because the Interaction Network is free to take as long as it wants to solve the problem and submit a result, lot of parameter tuning is necessary to get the reward structure just right:
\begin{itemize}
	\item At the end of each task, the temporal-difference learning should be disrupted, so that the reward depends only on the external reward for that iteration, and ignores the expected reward of the next iteration. This emulates episodic tasks without breaking the continuous nature of the IN. If you do not do this, the IN may learn to submit many poor solutions in rapid succession, instead of a single good one that would take multiple iterations to calculate.
	\item Failing to submit a result for too many iterations must be punished, or the IN may get stuck in an endless loop of procrastination and never submit a result. Picking the right number of iterations that we allow the IN to take is easy for this example task, since we know how many iterations the optimal solution will take. However, on novel tasks with unknown solutions, this becomes a hyperparameter that may be difficult to tune. Err too much in one direction and the IN wastes time. Err too much in the other direction and the IN is unable to solve the problem in the allotted time at all.
	\item Submitting a wrong result must be punished, but not too much: If the expected reward becomes negative at any point during the learning process, the IN may fall into a local optimum where it delays submitting a result as long as possible without triggering the punishment for taking too long. While the IN can break out of these loops on its own, this can take a very long time, since these loops are by their very nature very sparse with reward signals.
\end{itemize}

\paragraph{Changes in the sequence length during the training process}

The difficulty of the task increases with the length of the sequence that the IN is required to solve. This has a number of consequences:
\begin{itemize}
	\item If the required sequence length is increased too quickly, there is a risk that the IN becomes unable to solve the task at all and effectively dies out, because it can no longer get any positive reinforcement except by chance. It is necessary to reduce the sequence length again if the IN fails too often in a row.
	\item If the required sequence length is decreased too easily, the IN can fall into an interesting pattern where it effectively fails the task on purpose in order to make the task simpler again. This is not the behavior I intended, but it actually gives a greater expected reward than the intended behavior would.
\end{itemize}

\paragraph{Exploration vs Exploitation}

As with all Reinforcement Learners, Interaction Networks have a tradeoff between exploration and exploitation.

However, Interaction Networks are more strongly affected by this issue than normal reinforcement learners:

The number of iterations to solve one task is variable. The higher the sequence length, the higher the number of steps. This means that the probability of taking random exploration steps needs to be scaled according to the situation.

The problem we are trying to solve in this experiment does not have any room for error, and requires following the path to the solution precisely. With a high required sequence length, even a low probability of taking a random exploration step is likely to disrupt the sequence, which makes it impossible to reliably learn longer sequence lengths.

\paragraph{The function to learn}

It appears that the complexity of the function we want to learn does not affect the training difficulty as much as one would assume.

What matters the most is: Does the PU need to be executed often? Does it need to be executed reliably? Does it need to be executed a fixed number of times, or a variable number of times?

All of these things need to be learned by the Control Unit.

In contrast, the complexity of the function calculated by the PU does not seem to influence the result as much.

\paragraph{Unstable reward propagation}

While the complexity of the PU is not as important as I expected, what does matter is how chaotic the PU acts if it is slightly off. Learning XOR to solve the parity task is almost impossible, because every time the difficulty is increased, even a slight inaccuracy in the function learned so far causes the task to fail. Since the Interaction Network does not know which part of the sequence was wrong, the gradients from the punishment get applied to the wrong steps, and everything unravels.

I expect without having tested it, that a highly complex PU, such as for example one that recognizes patterns in images, would ironically be easier to integrate into an Interaction Network than learning the parity task. Learning the parity task requires many steps, which means many possible sources of error and an unstable learning process. Integrating a complex object recognition requires only a single PU that is executed once, which is less likely to lead to an unstable learning process.

\begin{figure}[H]
	\centering
	\fbox{
		\includegraphics[width=0.5\columnwidth]{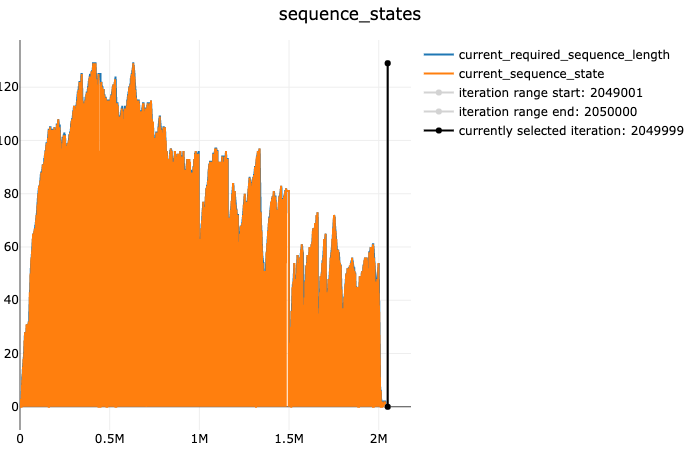}
	}
	\captionsetup{singlelinecheck=off}
	\caption[list=off]{This graph shows the maximum sequence length that was achieved by the IN. The function to predict was fairly simple, and it even had training wheels on the CU for 2,000,000 iterations. Despite this, the training process was unstable and even deteriorated after reaching a peak at around iteration 500,000. When the training wheels are taken off, it collapses completely.
	
	This seems to have two main reasons:
	
	\begin{itemize}
		\item The horizon of backpropagating the gradient through the PU's is not long enough. The first time it reaches a new length and makes a mistake, this mistake propagates back and destabilizes the entire chain. Learning such long-term dependencies is hard.
		\item The output of the learned function is supposed to be either 0 or 1, but during the training process it can also take on intermediate values. The longer the sequence, the more the inaccuracies add up, and the further the outputs get from 0 or 1. This is an interesting problem, because it has a solution that generalizes to other problems as well: If the intermediate results are rounded to 0 or 1, training becomes more stable. To make this work, there would need to be a mechanism (like a pretrained, more generic PU) that takes the best-guess of one PU and transforms it to better fit previously encountered values of that Node. It is interesting to note that humans do the same thing when performing explicit reasoning: We frequently pattern-match to things we are familiar with, because it is easier to reason about things you have encountered before.
	\end{itemize}
	}
	\label{fig:paper_1_rfa_unstable}
\end{figure}

See Figure \ref{fig:paper_1_rfa_unstable} for an example of unstable reward propagation.

\paragraph{Miscellaneous}

\begin{itemize}
	\item I tried giving the CU a memory of its own, like a recurrent neural network. Adding this memory makes it more flexible, since it no longer needs to rely on the PU's to change its inputs now. However, it also sometimes makes it slower to converge because its memory is one more thing that needs to be optimized.
	\item I have experimented with different types of meta-information for the CU to take as input. For example, indicator variables about which action was last chosen, and if that action was the highest rated one, or was picked because of random exploration. Check the code for a list of all of these variables. Adding more of this metadata usually increased performance, but it also sometimes slowed the learning process because many of the variables are irrelevant for any given task, and the IN has to learn to ignore them. The more complex the task, the more likely these values are to help, since it only takes a small, constant amount of time to learn to ignore irrelevant information, but a useful input can make the learning process much easier.
	\item I have experimented with different ways of backpropagating the gradients. There are several different choices for selecting the horizon for backpropagation, for scaling gradients that arrive at a network from different sources, and for accumulating gradients and applying them all in a batch. These experiments lead to differences in performance, but I did not notice a simple pattern. Tuning this is a future avenue of research.
\end{itemize}

\subsubsection{Summary}

Learning long sequences of actions is substantially more difficult than picking between alternative actions.

There are a lot of unusual pitfalls that are not encountered in other types of neural networks.

New methods will need to be developed to make Interaction Networks practical.

\section{Future Work}

There are a large number of possibilities to explore in the future. These are the key points that I consider most interesting.

\subsection{Explore the double-optimization problem}

The core problem with training Interaction Networks seems to be that two separate objectives need to be trained at the same time, and they interfere with each other. Investigating this in more detail could yield valuable, actionable insights.

The core of the problem is that the IN has to train each individual PU for its respective task, but at the same time also has to train the CU to pick which PU should be applied to which task in the first place.

The CU decides the training data of the PU, but the PU's performance on its data influences the CU. This can cause destructive interference, and destabilize the learning process.

The CU sometimes needs to keep trying the same sequence of actions to solve a problem even without a reward, because the PU's involved need to be trained for it first. But it also needs to know when using those PU's is just not the right approach at all and is a waste of time.

Another important aspect of this problem that hasn't been mentioned yet is that destructive interference actually outweighs useful learning processes: When a PU is already well-trained for a specific task, and the CU mistakenly uses it for a different task, then the gradient the PU receives from this will likely be higher than the gradient it receives on the task it already knows. This is counterproductive, because it means that a single wrongful use of a PU causes more damage to it than a correct use improves it. To prevent this, there should be a mechanism that alters the strength of a gradient so as to ensure that a PU will not unlearn its existing task when it is used experimentally.

I expect that the most effective way to make the double-optimization problem tractable is through Curriculum Learning.

When humans teach children to perform a task, we always start off with simple tasks. We literally have a saying for it: "You have to learn to walk before you can run."

I believe that this is no coincidence, and expect that the intelligent design of curricula will play an important part in AI development in general, and Interaction Networks in particular.

\subsection{Exploration vs Exploitation, Creativity vs Concentration}

The trade-off between exploration and exploitation is always an important question in Reinforcement Learning.

Normal Reinforcement Learning algorithms have to find a balance between exploration and exploitation when looking for rewards in the state space. The problem is slightly different for Interaction Networks: The state space explored by the CU is not the state of a game, but the mental state of the Interaction Network itself.

Because of this difference, using the words exploration and exploitation may be slightly misleading. Instead, 'creativity' and 'concentration' may be more apt descriptors.

It is possible to have an Interaction Network that emulates a DQN, where the DQN has a trade-off between exploration and exploitation of the environment. Independent of this, the CU itself has its own trade-off, between creativity and concentration.

If the Control Unit 'concentrates', it always uses the Processing Unit that is most likely to achieve a goal, but never explores alternative ideas. If the Control Unit is 'creative', it frequently uses Processing Units in previously unseen ways and may discover new ways to solve problems.

Learning a good trade-off for the CU is likely going to be very difficult because of the double-optimization problem mentioned above.

\subsection{Curiosity}

Curiosity is an inherent reward mechanism that allows an Interaction Network to learn even in the absence of effective external reward signals.

Curiosity can be implemented by defining a measure of uncertainty and rewarding the model if it learns to reduce its uncertainty. This has been used with great success in reinforcement learning\cite{burda2018largescale}.

There are multiple ways to implement a reduction of uncertainty:

\begin{itemize}
	\item Reduce uncertainty about the next state of the environment. This is what existing works in Reinforcement Learning tend to use. An Interaction Networks can use this as well.
	\item Reduce uncertainty about the Interaction Network's own internal state, as represented by its Nodes. The Nodes of the IN are likely to contain processed data derived from the Environment. Previous experiments on Reinforcement Learners have shown that curiosity is more effective if it is applied to an abstraction of the game state, instead of the game state itself. This suggests that predicting selected parts of the internal state of the IN would be more useful as a drive than predicting the raw Environment data.
	\item Reduce uncertainty about your own expected reward. This works by changing the CU so that it returns a probability distribution over expected rewards for each action, instead of only an expected reward. The curiosity drive rewards the network whenever it learns something that narrows these probability distributions. This approach seems most promising to me, since we optimize for exactly what we want: We learn to explore those actions in more detail about whose usefulness we are currently uncertain. This is more goal-directed than trying to minimize uncertainty in the state space of the game. Finding an effective formula to learn probability distributions over expected rewards is not trivial, but it may be worth the effort.
\end{itemize}

\subsection{Scenario Mechanisms}

Scenario Mechanisms are the Interaction Network's equivalent to Experience Replay in DQN's, but they are more general.

The basic idea is this: Instead of letting the Interaction Network operate as normal, cut off its connection to the outside world. Then replace the values in some Nodes with different values, and (optionally) alter the behavior of the CU and Environments. In this way, it becomes possible to simulate a number of different scenarios, each of which is useful for a different purpose.

\begin{itemize}
	\item Replay one iteration of the CU, using Node values that were previously used. This is equivalent to Experience Replay in DQN's.
	\item Replay individual PU executions that caused the Environments to apply loss functions to the Nodes. This allows us to train PU's more quickly. It simulates the way FFN's are normally trained outside of Interaction Networks: By repeatedly applying training samples to them until the network converges.
	\item Replay a short sequence of actions that previously lead to a reward. While doing so, ignore it if the CU now suggests different actions to take. This ensures that a useful sequence of actions that was found only once through random exploration gets reinforced, and is not just discarded. Finding a good way to determine sequences of actions that belong together will be a challenge. I expect that curiosity can help with this.
	\item Test what-ifs: As part of the creativity vs concentration trade-off, systematically test if executing some PU's would have created useful context data that would have helped the CU make better predictions. If the answer is positive, train the CU to use those PU's in similar situations in the future. If not, undo all changes to avoid introducing noise and disrupting the network unduly.
\end{itemize}

Many of these mechanisms also come with drawbacks, because they tend to bias the training process. It will be interesting to test if the benefits outweigh the costs in practice.

For some types of Scenario Mechanism, it would also make sense to allow the Control Unit to trigger the Scenario Mechanism deliberately, with an action. The reward for doing so would be based on the reduction of uncertainty that was achieved by running the Scenario Mechanism.

\subsection{Dynamic Network Modifications}

The Interaction Networks shown so far have used a static architecture, with a fixed number of Nodes and Processing Units.

As mentioned before, it would not present a major difficulty to add new Nodes or PU's to an existing IN. It just requires a small change to the CU.

The ability to generate new Nodes and PU's as needed could be very valuable. It would essentially allow the IN to perform the equivalent of Neural Architecture Search on its own internals.

The question is how to decide when it makes sense to generate new Nodes and PU's. Intuitively, it makes sense to generate new PU's and Nodes whenever existing ones are unable to solve their tasks effectively. Conversely, if a Node or PU ends up not being used after a while, it may be deleted again.

Formalizing this is likely going to be a very difficult task, but the ability to generate new neurons (Nodes) and connections between neurons (PU's) is an important feature of biological brains\cite{van2002functional}, so it may be worth exploring in detail.

\subsection{Compiling of gradients}

The gradients used for training in an Interaction Network are much more complex than in normal neural networks.

The CU receives its training loss through a completely different function than the PU's do, but it is possible for each of these rewards to be backpropagated to the same network weights. This makes it necessary to scale gradients carefully, so that one type of gradient does not outweigh the other.

Much work has been done in the area of gradient optimization\cite{ruder2016overview}. Unfortunately, this research generally assumes that the training samples of the network are drawn from a single distribution, which does not change much over time.

This assumption does not hold for Interaction Networks, both because gradients can backpropagate to a network through different routes, and because the nature of the task a PU is trained on may change if the CU decides to use it differently.

Further research is required to determine in how far this poses a problem, and how to fix it.

\subsection{Generalizations}

The way the Interaction Network works may be generalized further.

At its core, the idea behind an IN is simply to put one ML algorithm (CU) in charge of other ML algorithms (PU's), while allowing the results of the latter to influence the former.

This could be implemented in other ways than the ones outlined in this paper. Some of these generalizations may be worth exploring:

\begin{itemize}
	\item Instead of giving the CU one action per PU, give it one action per Node. Executing that action means that all PU's that feed into that Node get executed (alternatively, all PU's that feed \emph{from} that Node). Each PU's result is evaluated, and the one with the highest rating is used as the output of the Node. This is more computationally expensive and adds an additional learning process for the quality evaluation. However, it would also make it much easier to compare competing approaches for solving a problem, and to train them all in parallel, since the CU will now need a smaller number of discrete actions to perform the same sequence of processing steps.
	\item Nodes can have several subtypes: Single vector storage, addressable memory system, linear combiner of multiple inputs. Virtually any kind of memory mechanism could be implemented. It is very likely that an effective Interaction Network will use a combination of many different types of Nodes for different purposes.
	\item In addition to trainable PU's, also use auxiliary PU's that are not trainable and simply shuffle data around from one Node to another. This adds only very little overhead, and having the ability to efficiently move information between Nodes without distorting it could be helpful e.g. for implementing memory mechanisms.
	\item Instead of waiting for externally applied gradients from Environments, it is possible to use the output of one PU as the target value of another PU. If used properly, this could allow the IN to train new PU's to predict what older, already established chains of PU's do, which is useful for creating shortcuts. Similarly, long chains of backpropagation can be avoided using synthetic gradients, which have been shown to improve performance in related areas.\cite{jaderberg2017decoupled}
	\item Run several IN's in parallel, while letting them share some Nodes and/or PU’s. This allows parallelization, since each IN can run largely independently and the IN's only need to interact when they use their shared elements. By applying a penalty to the use of those elements, the networks could be taught to only perform expensive communications when necessary, reducing computational costs. Furthermore, the possibility of letting IN’s outsource tasks to other IN’s and being able to give each other rewards opens an even larger avenue of research: We would effectively be building a collection of independent but cooperating agents. This seems in many ways analogous to Marvin Minsky's idea of the Society of Mind\cite{minsky1988society}.
\end{itemize}

\section{Conclusion}

We have seen in thought experiments that Interaction Networks can theoretically improve many types of contemporary neural networks.

Experiments have shown that the concept is sound. However, they also reveal a number of challenges that do not appear in conventional neural networks, which make Interaction Networks very hard to train.

Finally, we investigated several potential directions for future work. Hopefully, these will make Interaction Networks practical, so that they can be used to enhance contemporary neural networks.

\section{Code}

Code is available upon request.

Implementing an Interaction Network is more complex than an ordinary neural network: Because of the unpredictable order of input-function-output sequences of PU's, backpropagating through an IN requires tracking the gradients explicitly, even though memories may in principle be stored for arbitrary amounts of time. Existing libraries such as PyTorch and TensorFlow are not designed for this eventuality, so it is necessary to extend them in order to implement Interaction Networks.

I have built a framework based on PyTorch. This framework can:
\begin{itemize}
	\item Automate experiments
	\item Run testcases
	\item Compare hyperparameters and model variants
	\item Generate statistics and interactive visualizations of the network
\end{itemize}

The graphics in this paper have been generated using that framework. The original graphics are interactive and can be inspected to see summary KPIs at a glance, and to track network behavior over time. They can be viewed in Jupyter Notebooks.

This interactive visualization is very useful for investigating how the behavior of the network changes over time.

At the moment, this project is a work in progress. If there is demand for it, I will spend some time to polish it and make it suitable for public release.

\bibliographystyle{unsrt}
\bibliography{bibliography}

\end{document}